\documentclass{article}

\PassOptionsToPackage{numbers, compress}{natbib}
\usepackage[preprint]{neurips_2022}
\usepackage{authblk}

\usepackage{wrapfig}
\usepackage[utf8]{inputenc} % allow utf-8 input
\usepackage[T1]{fontenc}    % use 8-bit T1 fonts
\usepackage{hyperref}       % hyperlinks
\usepackage{url}            % simple URL typesetting
\usepackage{booktabs}       % professional-quality tables
\usepackage{amsfonts}       % blackboard math symbols
\usepackage{nicefrac}       % compact symbols for 1/2, etc.
\usepackage{microtype}      % microtypography
\usepackage[table,xcdraw]{xcolor}

\usepackage{amsmath}
\usepackage{graphicx}
\usepackage{tikz}
\usepackage{pgfplots}
\usetikzlibrary{shapes.geometric}

\usepackage{multirow}
\title{Active Few-Shot Classification:\\a New Paradigm for Data-Scarce Learning Settings}

\author{%
\textit{Aymane Abdali}$^{12}$ \quad \textit{Vincent Gripon}$^{1}$ \quad \textit{Lucas Drumetz}$^1$ \quad \textit{Bartosz Boguslawski}$^2$ \\

$^1$IMT Atlantique Electronics Dept. \\ \quad $^2$Schneider Electric\\
}

\begin{document}

\maketitle

\begin{abstract}
  We consider a novel formulation of the problem of Active Few-Shot Classification (AFSC) where the objective is to classify a small, initially unlabeled, dataset given a very restrained labeling budget. This problem can be seen as a rival paradigm to classical Transductive Few-Shot Classification (TFSC), as both these approaches are applicable in similar conditions. We first propose a methodology that combines statistical inference, and an original two-tier active learning strategy that fits well into this framework.  We then adapt several standard vision benchmarks from the field of TFSC. Our experiments show the potential benefits of AFSC can be substantial, with gains in average weighted accuracy of up to 10\% compared to state-of-the-art TFSC methods for the same labeling budget. We believe this new paradigm could lead to new developments and standards in data-scarce learning settings.
\end{abstract}

\section{Introduction}

We consider the problem of learning to classify a small set of data samples with no given training set but a --very-- limited labeling budget instead. Such a problem is likely to arise in contexts where data samples are available yet their labeling is costly. These contexts are often encountered in real-world industrial applications. In our work we focus on the case of visual inputs but the rationale would be applicable to other domains as well, which is left as future work.

When the total number of data samples is large, Self-Supervised Learning (SSL) has recently emerged as a promising solution, showing the ability to reach high accuracy with a very limited number of labeled samples~\cite{caron2020unsupervised,chen2020simple,amrani2021self,boney2017semi}, yet requiring a large number of total samples.

When the number of data samples is reduced, the field of Transductive Few-Shot Classification (TFSC) comprises many solutions that can similarly achieve high accuracy. The rationale consists in using a large generic dataset to train efficient feature extractors, hoping they can generate meaningful such features with respect to the considered Few-Shot task. In the context of TFSC benchmarks, the labeled samples of the Few-Shot task are usually considered to be uniformly drawn at random.

The problem of selecting which samples to label in a classification task is not new. Indeed, the field of Active Learning has proposed many solutions to cope with this issue~\cite{cohn1996active,nguyen2004active,muller2022active,yuan2020cold}. More details are provided in Section~\ref{rw}. Yet, adapting active learning to the low-data regime, especially for vision, is under-explored and we believe it is relevant and meaningful to many real-world applications.

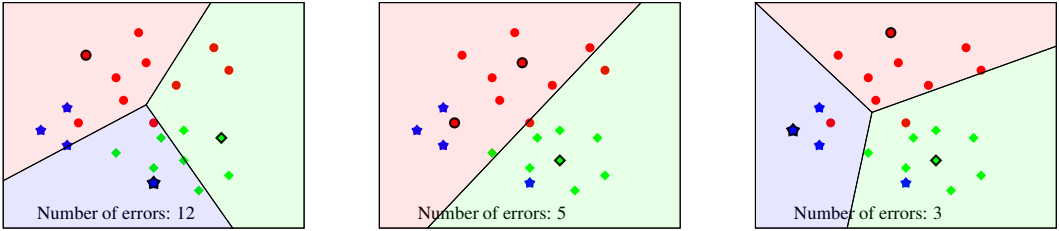
\begin{figure}[!h]
    \centering
    \begin{tikzpicture}{
    \begin{scope}[xshift=0cm]
    \draw
    (0,0) rectangle (4cm,3cm);
    \tikzset{class1/.style={circle, fill=red, inner sep=0pt, minimum width=0.12cm}}
    \tikzset{class2/.style={star, fill=blue, inner sep=0pt, minimum size=0.16cm}}
    \tikzset{class3/.style={diamond, fill=green, inner sep=0pt, minimum size=0.14cm}}
    \node[scale=0.7] at (1.5,0.2) {Number of errors: 12};
    \node [class1] at (1, 1.4) {};
    \node [class1] at (1.6, 1.7) {};
    \node [class1] at (1.5, 2) {};
    \node [class1] at (1.1, 2.3) {};
    \node [class1] at (1.9, 2.2) {};
    \node [class1] at (1.8, 2.6) {};
    \node [class1] at (3, 2.1) {};
    \node [class1] at (2.8, 2.4) {};
    \node [class1] at (2.3, 1.9) {};
    \node [class1] at (2, 1.4) {};
    \node [class2] at (0.5,1.3) {};
    \node [class2] at (0.85,1.1) {};
    \node [class2] at (0.85,1.6) {};
    \node [class3] at (2.1,1.2) {};
    \node [class3] at (1.5,1) {};
    \node [class3] at (2.4,1.3) {};
    \node [class3] at (2.9,1.2) {};
    \node [class3] at (2,0.8) {};
    \node [class3] at (2,0.6) {};
    \node [class3] at (2.6,0.5) {};
    \node [class3] at (2.4,0.9) {};
    \node [class3] at (3,0.7) {};
    \node[class1, draw=black, thick] at (1.1,2.3) {};
    \node[class2, draw=black, thick] at (2, 0.6) {};
    \node[class3, draw=black, thick] at (2.9,1.2) {};

    \draw [draw = black, fill=blue,fill opacity=0.1] (1.9,1.64) -- (0,0.63) -- (0,0) -- (3.05,0) -- cycle;
    \draw [draw = black, fill=red,fill opacity=0.1] (1.9,1.64) -- (0,0.63) -- (0,3) -- (2.76,3) -- cycle;
    \draw [draw = black, fill=green,fill opacity=0.1] (1.9,1.64) -- (3.05,0)-- (4,0) -- (4,3) --(2.76,3) -- cycle;
    %\draw
    %(0,1) edge (4,1.5);
    \end{scope}
        \begin{scope}[xshift=5cm]
    \draw
    (0,0) rectangle (4cm,3cm);
    \tikzset{class1/.style={circle, fill=red, inner sep=0pt, minimum width=0.12cm}}
    \tikzset{class2/.style={star, fill=blue, inner sep=0pt, minimum size=0.16cm}}
    \tikzset{class3/.style={diamond, fill=green, inner sep=0pt, minimum size=0.14cm}}

    \node[scale=0.7] at (1.5,0.2) {Number of errors: 5};
    \node [class1] at (1, 1.4) {};
    \node [class1] at (1.6, 1.7) {};
    \node [class1] at (1.5, 2) {};
    \node [class1] at (1.1, 2.3) {};
    \node [class1] at (1.9, 2.2) {};
    \node [class1] at (1.8, 2.6) {};
    \node [class1] at (3, 2.1) {};
    \node [class1] at (2.8, 2.4) {};
    \node [class1] at (2.3, 1.9) {};
    \node [class1] at (2, 1.4) {};
    \node [class2] at (0.5,1.3) {};
    \node [class2] at (0.85,1.1) {};
    \node [class2] at (0.85,1.6) {};
    \node [class3] at (2.1,1.2) {};
    \node [class3] at (1.5,1) {};
    \node [class3] at (2.4,1.3) {};
    \node [class3] at (2.9,1.2) {};
    \node [class3] at (2,0.8) {};
    \node [class2] at (2,0.6) {};
    \node [class3] at (2.6,0.5) {};
    \node [class3] at (2.4,0.9) {};
    \node [class3] at (3,0.7) {};
    \node[class1, draw=black, thick] at (1,1.4) {};
    \node[class1, draw=black, thick] at (1.9, 2.2) {};
    \node[class3, draw=black, thick] at (2.4,0.9) {};
    \draw [draw = black, fill=green,fill opacity=0.1] (0.64,0) -- (3.48,3) -- (4,3) --(4,0) -- cycle;
    \draw [draw = black, fill=red,fill opacity=0.1] (0,0) --(0.64,0) -- (3.48,3) -- (0,3) -- cycle;

    %\draw
    %(0,1) edge (4,1.5);
    \end{scope}
        \begin{scope}[xshift=10cm]
    \draw
    (0,0) rectangle (4cm,3cm);
    \tikzset{class1/.style={circle, fill=red, inner sep=0pt, minimum width=0.12cm}}
    \tikzset{class2/.style={star, fill=blue, inner sep=0pt, minimum size=0.16cm}}
    \tikzset{class3/.style={diamond, fill=green, inner sep=0pt, minimum size=0.14cm}}

    \node[scale=0.7] at (1.5,0.2) {Number of errors: 3};
    \node [class1] at (1, 1.4) {};
    \node [class1] at (1.6, 1.7) {};
    \node [class1] at (1.5, 2) {};
    \node [class1] at (1.1, 2.3) {};
    \node [class1] at (1.9, 2.2) {};
    \node [class1] at (1.8, 2.6) {};
    \node [class1] at (3, 2.1) {};
    \node [class1] at (2.8, 2.4) {};
    \node [class1] at (2.3, 1.9) {};
    \node [class1] at (2, 1.4) {};
    \node [class2] at (0.5,1.3) {};
    \node [class2] at (0.85,1.1) {};
    \node [class2] at (0.85,1.6) {};
    \node [class3] at (2.1,1.2) {};
    \node [class3] at (1.5,1) {};
    \node [class3] at (2.4,1.3) {};
    \node [class3] at (2.9,1.2) {};
    \node [class3] at (2,0.8) {};
    \node [class2] at (2,0.6) {};
    \node [class3] at (2.6,0.5) {};
    \node [class3] at (2.4,0.9) {};
    \node [class3] at (3,0.7) {};
    \node[class2, draw=black, thick] at (0.5,1.3) {};
    \node[class1, draw=black, thick] at (1.8, 2.6) {};
    \node[class3, draw=black, thick] at (2.4,0.9) {};

    \draw [draw = black, fill=red,fill opacity=0.1] (0,3) -- (0.1,3) -- (4,3) -- (4,2.42) --(1.55,1.54) -- cycle;
    \draw [draw = black, fill=blue,fill opacity=0.1] (0,3) -- (0,0) -- (1.22,0) --(1.55,1.54) -- cycle;
    \draw [draw = black, fill=green,fill opacity=0.1] (1.55,1.54) -- (1.22,0) --(4,0)-- (4,2.42) -- cycle;
    %\draw
    %(0,1) edge (4,1.5);
    \end{scope}
    }
    \end{tikzpicture}
    \caption{Illustration of the interest of the Active Few-Shot Classification formulation as compared to classical Transductive Few-Shot Classification. In the three rectangles, we depict the distribution of samples in three possible classes, represented as colors. On the left figure, the labeled samples, marked in bold, have been randomly drawn. On the center and on the right, the distribution of samples in each class have been estimated using an Expectation Maximization method. On the center, we select the samples that are the closest to the estimated Gaussian centers, whereas on the right we use the proposed criterion. Below the figures are indicated the performance of a simple nearest-neighbor classifier based on the labeled samples. We also depict the decision boundaries with the straight lines.}
    \label{fig:my_label}
\end{figure}

In this paper, we thus introduce a formulation of the above-mentioned problem of ``Active Few-Shot Classification'' (AFSC). We propose a methodology combining statistical inference with well-motivated heuristics, and then introduce benchmarks adapting the ones from the field of TFSC. Using them, we demonstrate the ability of the proposed framework to reach new levels in accuracy compared to classical solutions in TFSC. Our main motivation is to show it is possible to achieve new levels of performance when looking at the data-scarce learning problem from this new angle, as illustrated in Figure~\ref{fig:my_label}.

{\bf Contributions: }
\begin{itemize}
\item We propose a formulation the Active Few-Shot Classification problem. Under this framework, we empirically show it is possible to reach very high accuracy compared to classical TFSC methods, we discuss these improvements in more details in the experiments. As we believe that many TFSC real-world problems could be revisited as active Few-Shot classification ones, these accuracy improvements enable new possible applications.

\item We introduce a theoretically motivated two-tier methodology that consists in inferring the distribution of samples among the classes, and selecting the samples that are the most probable in each class to be labeled first, then the least probable ones. We compare this approach with classical ones from the Active Learning literature and show its ability to reach better accuracy in all proposed benchmarks.

\item We propose vision benchmarks that are adapted from the TFSC community on which we conduct the experiments and make our code available for reproducibility.
\end{itemize}

\section{Problem Statement}
\label{ps}
%\subsection{Problem statement}

Following the classical setting in Few-Shot classification, we suppose we are given a base dataset comprising possibly many classes and labeled samples in each class. This dataset can be used to train a generic feature extractor $f_{\theta}$. Next, we are given a small dataset $\mathcal{N}:= \left\{ \mathbf{x}_i,y_i \right\}^{N}_{i=1}$ made of $N$ samples that belong to $K$ novel classes. These classes are disjoint from those of the base dataset, and the samples can be arbitrarily distributed among classes. The Active Few-Shot Classification task consists in predicting the class of each sample in $\mathcal{N}$ after unveiling the label of at most $\ell \leq N$ samples. Performance is measured with weighted accuracy on all samples, including labeled ones.

For commodity, we refer to this problem as a $K$-way, $\ell$-labels, $N$-samples task. The samples are equivalent to the query set in standard TFSC setting whereas labels are equivalent to the total number of shots~\cite{dhillon2019baseline,hou2019cross,boudiaf2020information,liu2018transductive}. The major difference lies in the fact that the labeled samples are chosen instead of being randomly drawn. This results in a varying number of labels per class, including possibly zero for some of them in the worst case.

%Most of the recent literature in the standard TFSC $K$-way, $N$-shots $Q$-Queries setting assumes the marginal label probability of the queries is known and uniform i.e that the queries contain an equal number of examples per class. Recent work has shown that most state-of-the-art methods \cite{} have significant drops in performance when this assumption does not hold, which is often the case in realistic scenarios~\cite{}. Therefore, we model the marginal probabilities of the classes as Dirichlet-distributed random variables for our sample set as in~\cite{}. Moreover, forcing our labels to be within the samples represents an even more realistic setting for several applications as we lose the guarantee to obtain a uniform distribution for labeled shots as well.

\section{Methodology}
\label{Methodo}

The proposed method relies on multiple steps. The first one consists in training a feature extractor to transform raw images into high dimensional feature vectors that are hopefully easier to discriminate. We perform several preprocessing steps on the obtained feature vectors, including normalization and graph smoothing. Next, we infer a probability distribution on all samples using an Expectation-Maximization (EM) methodology. Finally, we propose a sequential active learning procedure that first picks the samples with the highest confidence for each class, then those with the lowest confidence. In the next paragraphs, we detail each of these steps.

\subsection{Feature extraction}
\label{fe}

Creating a feature extractor that can provide meaningful representations for samples outside the class domain it was initially trained for is an important problem in machine learning, as discussed in Section~\ref{rw}. It is in particular a very common step in the field of Few-Shot learning, where elaborate methods can significantly boost the quality of features. For example, the classical ResNet12 architecture originally proposed in~\cite{chen2019closer} was able to reach 57\% average accuracy on 5-way 1-shot problems generated from the mini-ImageNet dataset, while the same architecture can reach more than 70\% accuracy with a few adaptations as stated in~\cite{bendou2022easy}.

Since training an efficient feature extractor is already a classical step in the literature of Few-Shot learning, we simply use off-the-shelf models in this paper.

%After the feature extraction we smoothen our data through a smoothing process inspired from the Simple Graph Convolution \cite{}.We do this by constructing a graph from our data and averaging locally each point with its $m$ closer neighbors to a certain order $\kappa$. This process has proven that it significantly increases performance for distance-based methods, but also showed a interesting boost for other types of methods that we consider.

\subsection{Preprocessing}
\label{prepro}
Following~\cite{wang2019simpleshot}, we perform a two-step normalization, where feature vectors are first centered (the mean of the resulting feature vectors is 0), and then projected onto the unit-sphere (the norm of each feature vector becomes 1). We denote the resulting feature vector $\mathbf{x}$ and the feature vectors matrix $\mathbf{X}$.

\textbf{Graph Smoothing:} To exploit the information contained in the unlabeled samples, we propose to adapt the methodology described in~\cite{wu2019simplifying}. First, we create a graph whose vertices are the samples and whose edges depend on the cosine similarity between the associated feature vectors. Namely, we retain for each vertex its $m$ most similar neighbors obtaining a binary graph whose adjacency matrix is denoted by $\mathbf{A}$. Then we compute $\mathbf{M} = (\beta \mathbf{I} + \mathbf{A})^{\kappa}$, where $\beta$ and $\kappa$ are hyperparameters. We transform the feature vectors $\mathbf{X}$ into $\mathbf{Z} = \mathbf{M}\mathbf{X}$. Note that this operation has the effect of mixing each coordinate of a feature vector with the same coordinate in similar vectors, resulting in a smoothing operation~\cite{shuman2013emerging}, with the benefit of lowering the impact of outliers.

\subsection{Statistical inference}

Let us denote $\mathbf{z}_i$ the $i$-th preprocessed feature vector corresponding to the $i$-th sample in $\mathcal{N}$. We consider a simple model where, for each class $k$ the distribution of the corresponding $\mathbf{z}_{/y=k}$ follows a multivariate Gaussian with mean $\mathbf{\mu}_k$ and variance $\mathbf{\Sigma}_k$. If we consider an isotropic distribution, $\mathbf{\Sigma}_k$ becomes $\sigma_k  \mathbf{I}$ where $\sigma_k \in \mathbb{R}^{+}$ and $\mathbf{I}$ is the identity matrix.

Under this hypothesis, we can have a probabilistically optimal classifier given good estimations of the means and variances. On the other hand, we can obtain estimations by assigning examples to clusters. Algorithms that rely on Expectation Maximization (EM) allow us to find local optima of data partitioning. We focus on the soft $K$-means algorithm in this paper for two reasons; the first one being that soft $K$-means methods do not utilize, explicitly nor implicitly, any prior about the marginal label distributions. This property makes it robust to samples class imbalance. The second reason pertains to the unsupervised nature of the algorithm, as we can compute a locally optimal clustering with zero labels, albeit of suboptimal quality. We then obtain an estimate of $p(\mathbf{z}_{/y=k})$.

\subsection{Proposed active selection}

After the statistical inference, we want to quantify the potential increase in performance that data points would yield if they were to be labeled. Classical solutions rely on heuristics~\cite{boney2017semi} that include:

\begin{itemize}
\item {\bf  Margin}: The margin criterion, for any $\mathbf{z}_i$, is the difference between the highest probability (i.e probability of the assigned cluster) and the second best one. The sample with the highest margin is considered the most confident.

\item {\bf K-medoid}: The medoid is the cluster element that is closest to its centroid. The sample with the lowest distance to its cluster centroid is the one with the highest confidence.

\end{itemize}

Instead of using these classical approaches, we propose a new criterion based on the ratio of log densities that we call \textbf{Log-probs Soft K-means Sampling (LSS)}. Namely, we consider that the sample that has the lowest {\bf log-probability ratio (\textit{lpr})} of belonging to its cluster $k$ rather than other clusters is the most confident one.

%The density of $\mathbf{z}_{/y=k}$ is $$p(\mathbf{z}_{/y=k})= \frac{1}{\sqrt{2\pi}{\sigma}_k}\ex \left(-\frac{ || \mathbf{z}-{\mu_k} ||^{2} }{2{\sigma_k}}\right).$$

The logarithm of the density of $\mathbf{z}_{/y=k}$ is $$\log p(\mathbf{z}_{/y=k})= \log \left(\frac{1}{\sqrt{2\pi}{\sigma_k}}\right) - \frac{1}{2{\sigma_k}}\| \mathbf{z}-{\mu_k} \|^{2} $$We define the \textit{lpr} as the ratio of the log-densities over the different label-features distributions:

$$ lpr_k(\mathbf{z}) =\frac{\log
p(\mathbf{z}_{/y=k})}{\sum\limits_{i=1}^{K}\log p(\mathbf{z}_{/y=i})} $$

For  $a_k = \log \left(\frac{1}{\sqrt{2\pi}{\sigma_k}}\right)$ and $b_k = \nicefrac{1}{2\sigma_k}$
the log-probability ratio of sample $\mathbf{z}$ for cluster $k$ is:
$$lpr_k(\mathbf{z}) =\frac{a_k - b_k|| \mathbf{z}-{\mu_k} ||^{2} }{\sum\limits_{i=1}^{K} \left(a_i - b_i|| \mathbf{z}-{\mu_i} ||^{2} \right)} $$

We estimate the mean and the standard deviation of the class-features distribution using the clustering 

$$ \hat{\mu_k} = \frac{\sum\limits_{\mathbf{z} \in c_k}{\mathbf{z}}}{|c_k|} \text{  and  } \hat{\sigma_k} = \sqrt{\frac{\sum\limits_{\mathbf{z} \in C_k}^{}(\mathbf{z}-\hat{\mu_k})^2}{|C_k|}}.$$

Using the logs of the densities instead of the densities themselves makes the calculations more robust to the estimations. Under the assumption of a constant standard deviation for all the classes $\sigma = \nicefrac{1}{\sqrt{2\pi}}$ the ratio becomes: $$ lpr_k(\mathbf{z}) =\frac{|| \mathbf{z}-{\mu_k} ||^{2} }{\sum\limits_{i=1}^{K}|| \mathbf{z}-{\mu_i} ||^{2}}.$$ A log-prob ratio close to 0 indicates a strong confidence that the sample belongs to the cluster, whereas a ratio close to 1 indicates a high uncertainty. 

%\hspace{10pt} - {\bf Log-probabilities ratio}: We propose this criterion as a variation of the K-medoid that is more aware of the other clusters. This criterion examines the element's distance to its centroid, while simultaneously looking in the direction that is farthest from the other cluster centers. This ensures that the selected sample, when minimizing this criterion and hence maximizing confidence, has lower chances of belonging to any of the other clusers's distributions (more on this in the following section). 

Once we have chosen the criterion, we compute the value for each sample and select which ones to label based on this value~\cite{gu2015active}. We use a naive sequential pipeline for the active selection. This pipeline comprises several \emph{rounds}. The first round consists in selecting $K$ samples to label that yield the highest confidence according to the chosen criterion (e.g \textit{lpr} close to 0), that is one label per cluster. In the remaining rounds, we label samples that are chosen as those that yield the lowest possible confidence (e.g \textit{lpr} close to 1). The reasoning behind this two-tier strategy is that we want the first labels to be strongly representative of the class with lower risks of being outliers. Then we refine the statistical inference using these additional priors and choose labels that would be the most uncertain. As a matter of fact, samples with the strongest confidence are likely to be correctly predicted, hence there is no need to manually label them.

Since for the first round, we do not have any labels, we use multiple random intializations for the soft $K$-means. They will each generate a different clustering $\pi_1,\pi_2,....\pi_{n}$. We then use the clustering that minimizes the sum of the distances of the samples to their closest centroids.

For the subsequent rounds, we run a soft $K$-means with centroids initializations that correspond to the labels means for each class i.e the centroid of the $i^{th}$ cluster is initialized as the mean of all the labels that we have from class $i$ (or arbitrarily initialized as zero if we do not have any labels for class $i$). This allows for a higher quality clustering that we can use for another round to select the next $K$ labels, or to classify the examples in the final round. In the next section, we compare our criterion with the classical ones in terms of average accuracy.

\section{Experiments}
\label{exp}

\subsection{Datasets}
\label{ds}
We adapt four standard benchmarks to AFSC: mini-ImageNet, tiered-ImageNet, CUB and FC-100. The details are as follows: 

\textbf{mini-ImageNet}: a subset of ImageNet \cite{russakovsky2015imagenet} that comprises 60,000 color images of size $84 \times  84$ pixels equally distributed over 100 classes. In all experiments, we use the standard split of 64 base-training, 16 validation and 20 test classes~\cite{wang2019simpleshot}. 
\textbf{tiered-ImageNet}: another larger subset of ImageNet that contains 608 classes and 779,165 color images of size $84 \times 84$ pixels. We use a split of 351 base-training, 97 validation and 160 test classes~\cite{wang2019simpleshot}. 
\textbf{CUB}: this benchmark contains 11788 images of size $84 \times 84$ pixels, with 200 classes. For CUB, we use a standard split of 100 base-training, 50 validation and 50 test classes, as in~\cite{chen2019closer}. 
\textbf{FC-100}: this benchmark is a subset of CIFAR-100 that is used for Few-Shot Learning. it contains 600 images of size $32\times 32$ for each of its 100 classes. We use a standard split of 60 base-training, 20 validation and 20 test classes~\cite{oreshkin2018tadam,dhillon2019baseline}.

We use a single NVIDIA GeForce RTX 3090 with 24GB of RAM for all our experiments and our code is available at \url{https://anonymous.4open.science/r/AFSC-F5E9/}.

\subsection{Benchmarking protocol}
\label{bp}
Our default settings are 5-ways $\ell$-labels, where $\ell=5,25$, with a sample set of size $75+\ell$.
Following~\cite{veilleux2021realistic}, the classes of the sample sets are randomly distributed following Dirichlet’s distribution with parameters $\alpha \mathbf{1}$, where $\mathbf{1}$ is the constant unit vector with $K$ coordinates and $\alpha = 2$. Unless specified otherwise, we evaluate the methods by computing the average score over 10,000 tasks~\cite{boudiaf2020information}. We chose to use the weighted accuracy (the average of recalls for each class) as the unweighted accuracy is not suited for extreme label imbalance. We also chose to compute this accuracy over all samples, including the labeled ones, for which we force the prediction to the value that we revealed.

For our experiments, we use two different feature extractors. The first one is a ResNet-18 with standard cross-entropy minimization on the base
classes and label smoothing. This feature extractor is trained for 90 epochs with a batch size of 256. The learning rate is initialized at 0.1 and divided by 10 at epochs 45 and 66 \cite{boudiaf2020information,veilleux2021realistic}. 
We use this feature extractor in Tables~\ref{tab:table1_soa}~and~\ref{soa_bis} to compare to state-of-the-art methods in the classical TFSC Setting ($N$-ways, $K$-shots, $Q$-queries \cite{veilleux2021realistic,snell2017prototypical,boudiaf2020information}).%We chose to reproduce our method in the configuration where those methods were evaluated to have a fair benchmark, we refer to this configuration as Equal labels per class in Table 1. We also applied our methods to the configurations that are most relevant to this paper (Label Sampling) cf. Table 1. 
The second feature extractor we use is the Ensembled Augmented Shots Y-shaped ResNet-12 from~\cite{bendou2022easy}: it consists of a concatenation of 3 ResNet-12 architectures of around 12 million parameters each. The training is performed over 5 cycles of 100 epochs each, with a cosine-annealing scheduler i.e a learning rate that decreases from $\nu_0$ to reach 0 at the end of the cycle. Namely, $\nu_0$ is initialized at 0.1 and decreased by 10\% for each cycle. This feature extractor allows to reach top-tier performance on classical TFSC problems, which is why we use it for the remainder of the experiments.

We use a grid search to fix the values of the hyperparameters on the validation set of mini-ImageNet and use the same hyperparameters throughout all of the experiments.

%We used this feature extractor to compare different active sampling criterions for the datasets with the random selection and the "best" selection cf. Table 2. We also use this feature extractor to measure the performance boost from random to active sampling in different settings e.g ( varying the number of samples, number of labels and for various $\alpha$-dirichlet values) cf. Figure 2.

\subsection{Advantages of AFSC compared to TFSC}

In the first experiment, we wanted to quantify the interest of using Active Few-Shot Classification when compared to the usual Transductive Few-Shot Classification (TFSC) approach. A difficulty in comparing raw performance is that in TFSC, a fixed number of labels per class is granted. This can be beneficial as it ensures a fair representation of the various classes in the considered task. In Tables~\ref{tab:table1_soa} and~\ref{soa_bis}, we report the performance of TFSC methods compared with AFSC. To better account for the added difficulty of potentially missing labels in some classes, we report results of the soft $K$-means statistical inference method under the same conditions as for the reported results (this is named ``Equal labels per class'') i.e we use the same feature extractor, same benchmark and same evaluation method. We then compare the performance when selecting labels at random and using the proposed active method. As we only had access to results from~\cite{veilleux2021realistic}, we report here the average unweighted accuracy instead of the weighted one. We also updated the values from~\cite{veilleux2021realistic} to reflect our measure of accuracy where labeled samples are accounted for.

\begin{table}
\centering
\setlength{\extrarowheight}{0pt}
\addtolength{\extrarowheight}{\aboverulesep}
\addtolength{\extrarowheight}{\belowrulesep}
\setlength{\aboverulesep}{0pt}
\setlength{\belowrulesep}{0pt}
\caption{Comparison of accuracy between state-of-the-art TFSC methods and the proposed AFSC one on mini-ImageNet and tiered-ImageNet. The same feature extractors are used in all experiments. We also indicate the performance of the statistical inference method (soft K-means) combined with the proposed preprocessing for completeness. We also include, for the accuracies that we computed, the 95\% confidence interval. All values are percentages.}
\label{tab:table1_soa}
\small
\begin{tabular}{llllll} 
\toprule
                                                                                                             &                       & \multicolumn{4}{c}{\textbf{mini-ImageNet}}                             \\ 
\midrule
\textbf{Shots configuration}                                                                                 & \textbf{method}      & 5-labels        & 25-labels       & 50-labels       & 100-labels       \\ 
\midrule
\multirow{6}{*}{\begin{tabular}[c]{@{}l@{}}\textbf{Equal labels per class}\\\textbf{ (random)}\end{tabular}} & PT-MAP \cite{hu2021leveraging}               & 62.6            & 75.3            & 81.3            & 70.4             \\
                                                                                                             & LaplacianShot \cite{ziko2020laplacian}         & 67.6            & 86.2            & 90.5            & 94               \\
                                                                                                             & BD-CSPN  \cite{liu2020prototype}                & 69.0            & 85.1            & 89.7            & 93.4             \\
                                                                                                             & TIM \cite{boudiaf2020information}                 & 69.3            & 84.5            & 89.4            & 93.2             \\
                                                                                                             & $\alpha$-TIM \cite{veilleux2021realistic}          & 69.4            & 86.9            & 91.6            & 94.8             \\
                                                                                                             & \textbf{soft K-means} & 70.7 $\pm$ 0.28 & 83.4 $\pm$ 0.16 & 87.7 $\pm$ 0.12 & 91.8 $\pm$ 0.09  \\ 
\midrule
\textbf{Random label sampling}                                                                               & \textbf{soft K-means} & 64.3 $\pm$ 0.29 & 84.7 $\pm$ 0.14 & 89.6 $\pm$ 0.09 & 93.3 $\pm$ 0.06  \\ 
\midrule
\rowcolor[rgb]{0.90,0.90,0.90} \textbf{Active label sampling}                                             & \textbf{LSS (ours)}   & 73.5 $\pm$ 0.21 & 87.7 $\pm$ 0.16 & 93.8 $\pm$ 0.11 & 96.8 $\pm$ 0.07  \\ 
\midrule
                                                                                                             &                       & \multicolumn{4}{c}{\textbf{tiered-ImageNet}}                           \\ 
\cmidrule(l){3-6}
                                                                                                             & \textbf{}             & 5-labels        & 25-labels       & 50-labels       & 100-labels       \\ 
\midrule
\multirow{6}{*}{\begin{tabular}[c]{@{}l@{}}\textbf{Equal labels per class}\\\textbf{ (random)}\end{tabular}} & PT-MAP \cite{hu2021leveraging}              & 63.6            & 77.5            & 83.1            & 88.6             \\
                                                                                                             & LaplacianShot \cite{ziko2020laplacian}         & 74.0            & 89.3            & 92.8            & 95.3             \\
                                                                                                             & BD-CSPN  \cite{liu2020prototype}                & 75.8            & 88.7            & 92.0            & 94.8             \\
                                                                                                             & TIM \cite{boudiaf2020information}                & 75.8            & 88.1            & 91.2            & 94.6             \\
                                                                                                             & $\alpha$-TIM \cite{veilleux2021realistic}          & 76              & 89.9            & 93.6            & 96.1             \\
                                                                                                             & \textbf{soft K-means} & 74.7 $\pm$ 0.29 & 84.8 $\pm$ 0.18 & 88.8 $\pm$ 0.14 & 92.6 $\pm$ 0.10  \\ 
\midrule
\textbf{Random label sampling}                                                                               & \textbf{soft K-means} & 66.9 $\pm$ 0.30 & 85.7 $\pm$ 0.16 & 90.0 $\pm$ 0.11 & 93.5 $\pm$ 0.07  \\ 
\midrule
\rowcolor[rgb]{0.90,0.90,0.90} \textbf{Active label sampling}                                             & \textbf{LSS (ours)}   & 76.1 $\pm$ 0.24 & 87.5 $\pm$ 0.17 & 92.2 $\pm$ 0.13 & 95.8 $\pm$ 0.09  \\
\midrule
\vspace{-1cm}
\end{tabular}
\end{table}

\begin{wraptable}{r}{8cm}
\caption{Same comparison as in Table~\ref{tab:table1_soa} on the CUB dataset. We only consider the 5-labels scenario for this dataset that only has 60 examples per class: for a Dirichlet distribution of the class labels and for sample size of 100 or more, we observe frequent occurrences of class distributions that largely surpass 60 examples for a given class. All values are percentages.}
\label{soa_bis}
\small
\begin{tabular}{llc}
\textbf{Shots configuration}                                                                                 & \textbf{method}    & 5-labels      \\ 
\cmidrule(lr){1-3}
\multirow{6}{*}{\begin{tabular}[c]{@{}l@{}}\textbf{Equal labels}\\\textbf{per class (random)}\end{tabular}} & PT-MAP \cite{hu2021leveraging}          & 67.3          \\
                                                                                                             & LaplacianShot \cite{ziko2020laplacian}       & 84.7          \\
                                                                                                             & BD-CSPN  \cite{liu2020prototype}            & 76.1          \\
                                                                                                             & TIM    \cite{boudiaf2020information}             & 76.4          \\
                                                                                                             & $\alpha$-TIM \cite{veilleux2021realistic}        & 77.2          \\
                                                                                                             & \textbf{soft K-means} & 75.9 $\pm$ 0.28   \\ 
\cmidrule(lr){1-3}
\textbf{Random sampling}                                                                               & \textbf{soft K-means} & 66.6 $\pm$ 0.29   \\ 
\cmidrule(lr){1-3}
\rowcolor[rgb]{0.9,0.9,0.9} \textbf{Active sampling}                                                                               & \textbf{LSS (ours)}        & 77.2 $\pm$ 0.21  \\
\cmidrule(lr){1-3}
\end{tabular}

\end{wraptable}

Results show that, across all datasets, We benefit from a large increase in accuracy when we opt for an active selection of labels rather than random. In Tables~\ref{tab:table1_soa} and~\ref{soa_bis}, we can clearly see that, in similar conditions, our active sampling makes it possible to obtain better scores compared to when we randomly pre-select equally distributed labels per class, which is the configuration that all the methods that we benchmark against were tested in. We also observe that, as the label number increases, the gap between random and active selection scores closes. This is partly due to the way of computing the accuracy in our formulation of the problem, as we include the labeled examples in the computation of the weighted accuracy. The other reason this gap is reduced is simply that for a larger pool of labeled samples, we are more likely to have interesting samples drawn randomly compared to when the pool is smaller. Another interesting effect in Table~\ref{tab:table1_soa} is that the Random label sampling configuration catches up to the Equal labels per class configuration as the number of labels increase and ultimately outperforms it, which highlights how drawing uniform labels from an uneven distribution could harm the accuracy.

\begin{figure}[h!]
\begin{center}
\begin{tikzpicture}
\definecolor{c1}{rgb}{0.67, 0.9, 0.93}
\definecolor{c2}{rgb}{0.0, 0.18, 0.39}
\tikzstyle{every node} = [draw, rectangle, minimum width=1em, minimum height=1em];
\node[fill=c1!58!c2] at (0em,9em) {};
\node[fill=c1!70!c2] at (1em,9em) {};
\node[fill=c1!72!c2] at (2em,9em) {};
\node[fill=c1!75!c2] at (3em,9em) {};
\node[fill=c1!77!c2] at (4em,9em) {};
\node[fill=c1!77!c2] at (5em,9em) {};
\node[fill=c1!79!c2] at (6em,9em) {};
\node[fill=c1!80!c2] at (7em,9em) {};
\node[fill=c1!79!c2] at (8em,9em) {};
\node[fill=c1!79!c2] at (9em,9em) {};
\node[fill=c1!37!c2] at (0em,8em) {};
\node[fill=c1!49!c2] at (1em,8em) {};
\node[fill=c1!50!c2] at (2em,8em) {};
\node[fill=c1!50!c2] at (3em,8em) {};
\node[fill=c1!48!c2] at (4em,8em) {};
\node[fill=c1!49!c2] at (5em,8em) {};
\node[fill=c1!50!c2] at (6em,8em) {};
\node[fill=c1!46!c2] at (7em,8em) {};
\node[fill=c1!44!c2] at (8em,8em) {};
\node[fill=c1!45!c2] at (9em,8em) {};
\node[fill=c1!18!c2] at (0em,7em) {};
\node[fill=c1!44!c2] at (1em,7em) {};
\node[fill=c1!44!c2] at (2em,7em) {};
\node[fill=c1!44!c2] at (3em,7em) {};
\node[fill=c1!42!c2] at (4em,7em) {};
\node[fill=c1!39!c2] at (5em,7em) {};
\node[fill=c1!37!c2] at (6em,7em) {};
\node[fill=c1!36!c2] at (7em,7em) {};
\node[fill=c1!35!c2] at (8em,7em) {};
\node[fill=c1!33!c2] at (9em,7em) {};
\node[fill=c1!0!c2] at (0em,6em) {};
\node[fill=c1!38!c2] at (1em,6em) {};
\node[fill=c1!44!c2] at (2em,6em) {};
\node[fill=c1!43!c2] at (3em,6em) {};
\node[fill=c1!40!c2] at (4em,6em) {};
\node[fill=c1!39!c2] at (5em,6em) {};
\node[fill=c1!37!c2] at (6em,6em) {};
\node[fill=c1!34!c2] at (7em,6em) {};
\node[fill=c1!31!c2] at (8em,6em) {};
\node[fill=c1!30!c2] at (9em,6em) {};
\node[fill=c1!0!c2] at (0em,5em) {};
\node[fill=c1!30!c2] at (1em,5em) {};
\node[fill=c1!40!c2] at (2em,5em) {};
\node[fill=c1!41!c2] at (3em,5em) {};
\node[fill=c1!41!c2] at (4em,5em) {};
\node[fill=c1!39!c2] at (5em,5em) {};
\node[fill=c1!37!c2] at (6em,5em) {};
\node[fill=c1!36!c2] at (7em,5em) {};
\node[fill=c1!34!c2] at (8em,5em) {};
\node[fill=c1!31!c2] at (9em,5em) {};
\node[fill=c1!0!c2] at (0em,4em) {};
\node[fill=c1!19!c2] at (1em,4em) {};
\node[fill=c1!37!c2] at (2em,4em) {};
\node[fill=c1!40!c2] at (3em,4em) {};
\node[fill=c1!41!c2] at (4em,4em) {};
\node[fill=c1!41!c2] at (5em,4em) {};
\node[fill=c1!38!c2] at (6em,4em) {};
\node[fill=c1!37!c2] at (7em,4em) {};
\node[fill=c1!34!c2] at (8em,4em) {};
\node[fill=c1!33!c2] at (9em,4em) {};
\node[fill=c1!0!c2] at (0em,3em) {};
\node[fill=c1!0!c2] at (1em,3em) {};
\node[fill=c1!25!c2] at (2em,3em) {};
\node[fill=c1!35!c2] at (3em,3em) {};
\node[fill=c1!40!c2] at (4em,3em) {};
\node[fill=c1!38!c2] at (5em,3em) {};
\node[fill=c1!38!c2] at (6em,3em) {};
\node[fill=c1!37!c2] at (7em,3em) {};
\node[fill=c1!36!c2] at (8em,3em) {};
\node[fill=c1!35!c2] at (9em,3em) {};
\node[fill=c1!0!c2] at (0em,2em) {};
\node[fill=c1!0!c2] at (1em,2em) {};
\node[fill=c1!12!c2] at (2em,2em) {};
\node[fill=c1!26!c2] at (3em,2em) {};
\node[fill=c1!33!c2] at (4em,2em) {};
\node[fill=c1!36!c2] at (5em,2em) {};
\node[fill=c1!37!c2] at (6em,2em) {};
\node[fill=c1!37!c2] at (7em,2em) {};
\node[fill=c1!38!c2] at (8em,2em) {};
\node[fill=c1!36!c2] at (9em,2em) {};
\node[fill=c1!0!c2] at (0em,1em) {};
\node[fill=c1!0!c2] at (1em,1em) {};
\node[fill=c1!0!c2] at (2em,1em) {};
\node[fill=c1!3!c2] at (3em,1em) {};
\node[fill=c1!17!c2] at (4em,1em) {};
\node[fill=c1!25!c2] at (5em,1em) {};
\node[fill=c1!29!c2] at (6em,1em) {};
\node[fill=c1!32!c2] at (7em,1em) {};
\node[fill=c1!34!c2] at (8em,1em) {};
\node[fill=c1!34!c2] at (9em,1em) {};
\node[fill=c1!0!c2] at (0em,0em) {};
\node[fill=c1!0!c2] at (1em,0em) {};
\node[fill=c1!0!c2] at (2em,0em) {};
\node[fill=c1!0!c2] at (3em,0em) {};
\node[fill=c1!0!c2] at (4em,0em) {};
\node[fill=c1!10!c2] at (5em,0em) {};
\node[fill=c1!18!c2] at (6em,0em) {};
\node[fill=c1!23!c2] at (7em,0em) {};
\node[fill=c1!27!c2] at (8em,0em) {};
\node[fill=c1!29!c2] at (9em,0em) {};

\tikzstyle{every node} = [];
\node at (-1em,0em) {\tiny{100}};
\node at (-1em,1em) {\tiny{75}};
\node at (-1em,2em) {\tiny{50}};
\node at (-1em,3em) {\tiny{40}};
\node[rotate=90] at (-2.5em,4em) {Number of labels};
\node at (4.5em,10.5em) {$\alpha$=1};
\node at (-1em,4em) {\tiny{30}};
\node at (-1em,5em) {\tiny{25}};
\node at (-1em,6em) {\tiny{20}};
\node at (-1em,7em) {\tiny{15}};
\node at (-1em,8em) {\tiny{10}};
\node at (-1em,9em) {\tiny{5}};
\node at (0em,-1em) {\tiny{20}};
\node at (1em,-1em) {\tiny{40}};
\node at (2em,-1em) {\tiny{60}};
\node at (3em,-1em) {\tiny{80}};
\node at (4em,-1em) {\tiny{100}};
\node at (5em,-1em) {\tiny{120}};
\node at (6em,-1em) {\tiny{140}};
\node at (7em,-1em) {\tiny{160}};
\node at (8em,-1em) {\tiny{180}};
\node at (9em,-1em) {\tiny{200}};

\begin{scope}[xshift=4cm]
\tikzstyle{every node} = [draw, rectangle, minimum width=1em, minimum height=1em];

\node[fill=c1!56!c2] at (0em,9em) {};
\node[fill=c1!78!c2] at (1em,9em) {};
\node[fill=c1!81!c2] at (2em,9em) {};
\node[fill=c1!86!c2] at (3em,9em) {};
\node[fill=c1!88!c2] at (4em,9em) {};
\node[fill=c1!89!c2] at (5em,9em) {};
\node[fill=c1!90!c2] at (6em,9em) {};
\node[fill=c1!90!c2] at (7em,9em) {};
\node[fill=c1!91!c2] at (8em,9em) {};
\node[fill=c1!91!c2] at (9em,9em) {};
\node[fill=c1!25!c2] at (0em,8em) {};
\node[fill=c1!36!c2] at (1em,8em) {};
\node[fill=c1!37!c2] at (2em,8em) {};
\node[fill=c1!38!c2] at (3em,8em) {};
\node[fill=c1!37!c2] at (4em,8em) {};
\node[fill=c1!37!c2] at (5em,8em) {};
\node[fill=c1!37!c2] at (6em,8em) {};
\node[fill=c1!35!c2] at (7em,8em) {};
\node[fill=c1!33!c2] at (8em,8em) {};
\node[fill=c1!34!c2] at (9em,8em) {};
\node[fill=c1!14!c2] at (0em,7em) {};
\node[fill=c1!29!c2] at (1em,7em) {};
\node[fill=c1!30!c2] at (2em,7em) {};
\node[fill=c1!26!c2] at (3em,7em) {};
\node[fill=c1!26!c2] at (4em,7em) {};
\node[fill=c1!23!c2] at (5em,7em) {};
\node[fill=c1!22!c2] at (6em,7em) {};
\node[fill=c1!21!c2] at (7em,7em) {};
\node[fill=c1!19!c2] at (8em,7em) {};
\node[fill=c1!19!c2] at (9em,7em) {};
\node[fill=c1!0!c2] at (0em,6em) {};
\node[fill=c1!27!c2] at (1em,6em) {};
\node[fill=c1!29!c2] at (2em,6em) {};
\node[fill=c1!26!c2] at (3em,6em) {};
\node[fill=c1!24!c2] at (4em,6em) {};
\node[fill=c1!22!c2] at (5em,6em) {};
\node[fill=c1!20!c2] at (6em,6em) {};
\node[fill=c1!18!c2] at (7em,6em) {};
\node[fill=c1!17!c2] at (8em,6em) {};
\node[fill=c1!15!c2] at (9em,6em) {};
\node[fill=c1!0!c2] at (0em,5em) {};
\node[fill=c1!24!c2] at (1em,5em) {};
\node[fill=c1!29!c2] at (2em,5em) {};
\node[fill=c1!27!c2] at (3em,5em) {};
\node[fill=c1!25!c2] at (4em,5em) {};
\node[fill=c1!23!c2] at (5em,5em) {};
\node[fill=c1!20!c2] at (6em,5em) {};
\node[fill=c1!19!c2] at (7em,5em) {};
\node[fill=c1!18!c2] at (8em,5em) {};
\node[fill=c1!15!c2] at (9em,5em) {};
\node[fill=c1!0!c2] at (0em,4em) {};
\node[fill=c1!16!c2] at (1em,4em) {};
\node[fill=c1!27!c2] at (2em,4em) {};
\node[fill=c1!27!c2] at (3em,4em) {};
\node[fill=c1!26!c2] at (4em,4em) {};
\node[fill=c1!24!c2] at (5em,4em) {};
\node[fill=c1!23!c2] at (6em,4em) {};
\node[fill=c1!21!c2] at (7em,4em) {};
\node[fill=c1!18!c2] at (8em,4em) {};
\node[fill=c1!17!c2] at (9em,4em) {};
\node[fill=c1!0!c2] at (0em,3em) {};
\node[fill=c1!0!c2] at (1em,3em) {};
\node[fill=c1!20!c2] at (2em,3em) {};
\node[fill=c1!26!c2] at (3em,3em) {};
\node[fill=c1!27!c2] at (4em,3em) {};
\node[fill=c1!26!c2] at (5em,3em) {};
\node[fill=c1!26!c2] at (6em,3em) {};
\node[fill=c1!25!c2] at (7em,3em) {};
\node[fill=c1!22!c2] at (8em,3em) {};
\node[fill=c1!21!c2] at (9em,3em) {};
\node[fill=c1!0!c2] at (0em,2em) {};
\node[fill=c1!0!c2] at (1em,2em) {};
\node[fill=c1!11!c2] at (2em,2em) {};
\node[fill=c1!21!c2] at (3em,2em) {};
\node[fill=c1!25!c2] at (4em,2em) {};
\node[fill=c1!26!c2] at (5em,2em) {};
\node[fill=c1!26!c2] at (6em,2em) {};
\node[fill=c1!25!c2] at (7em,2em) {};
\node[fill=c1!25!c2] at (8em,2em) {};
\node[fill=c1!24!c2] at (9em,2em) {};
\node[fill=c1!0!c2] at (0em,1em) {};
\node[fill=c1!0!c2] at (1em,1em) {};
\node[fill=c1!0!c2] at (2em,1em) {};
\node[fill=c1!3!c2] at (3em,1em) {};
\node[fill=c1!14!c2] at (4em,1em) {};
\node[fill=c1!20!c2] at (5em,1em) {};
\node[fill=c1!22!c2] at (6em,1em) {};
\node[fill=c1!24!c2] at (7em,1em) {};
\node[fill=c1!25!c2] at (8em,1em) {};
\node[fill=c1!25!c2] at (9em,1em) {};
\node[fill=c1!0!c2] at (0em,0em) {};
\node[fill=c1!0!c2] at (1em,0em) {};
\node[fill=c1!0!c2] at (2em,0em) {};
\node[fill=c1!0!c2] at (3em,0em) {};
\node[fill=c1!0!c2] at (4em,0em) {};
\node[fill=c1!9!c2] at (5em,0em) {};
\node[fill=c1!15!c2] at (6em,0em) {};
\node[fill=c1!19!c2] at (7em,0em) {};
\node[fill=c1!21!c2] at (8em,0em) {};
\node[fill=c1!22!c2] at (9em,0em) {};
\tikzstyle{every node} = [];
\node at (4.5em,10.5em) {$\alpha$=2};
\node at (0em,-1em) {\tiny{20}};
\node at (1em,-1em) {\tiny{40}};
\node at (2em,-1em) {\tiny{60}};
\node at (3em,-1em) {\tiny{80}};
\node at (4em,-1em) {\tiny{100}};
\node at (5em,-1em) {\tiny{120}};
\node at (6em,-1em) {\tiny{140}};
\node at (7em,-1em) {\tiny{160}};
\node at (8em,-1em) {\tiny{180}};
\node at (9em,-1em) {\tiny{200}};
\node at (4em,-2.5em) {Number of samples};
\end{scope}

\begin{scope}[xshift=8cm]
\tikzstyle{every node} = [draw, rectangle, minimum width=1em, minimum height=1em];

\node[fill=c1!55!c2] at (0em,9em) {};
\node[fill=c1!82!c2] at (1em,9em) {};
\node[fill=c1!90!c2] at (2em,9em) {};
\node[fill=c1!93!c2] at (3em,9em) {};
\node[fill=c1!94!c2] at (4em,9em) {};
\node[fill=c1!96!c2] at (5em,9em) {};
\node[fill=c1!95!c2] at (6em,9em) {};
\node[fill=c1!100!c2] at (7em,9em) {};
\node[fill=c1!98!c2] at (8em,9em) {};
\node[fill=c1!98!c2] at (9em,9em) {};
\node[fill=c1!12!c2] at (0em,8em) {};
\node[fill=c1!25!c2] at (1em,8em) {};
\node[fill=c1!27!c2] at (2em,8em) {};
\node[fill=c1!30!c2] at (3em,8em) {};
\node[fill=c1!29!c2] at (4em,8em) {};
\node[fill=c1!29!c2] at (5em,8em) {};
\node[fill=c1!27!c2] at (6em,8em) {};
\node[fill=c1!28!c2] at (7em,8em) {};
\node[fill=c1!26!c2] at (8em,8em) {};
\node[fill=c1!27!c2] at (9em,8em) {};
\node[fill=c1!7!c2] at (0em,7em) {};
\node[fill=c1!18!c2] at (1em,7em) {};
\node[fill=c1!18!c2] at (2em,7em) {};
\node[fill=c1!16!c2] at (3em,7em) {};
\node[fill=c1!15!c2] at (4em,7em) {};
\node[fill=c1!15!c2] at (5em,7em) {};
\node[fill=c1!13!c2] at (6em,7em) {};
\node[fill=c1!13!c2] at (7em,7em) {};
\node[fill=c1!12!c2] at (8em,7em) {};
\node[fill=c1!9!c2] at (9em,7em) {};
\node[fill=c1!0!c2] at (0em,6em) {};
\node[fill=c1!16!c2] at (1em,6em) {};
\node[fill=c1!17!c2] at (2em,6em) {};
\node[fill=c1!16!c2] at (3em,6em) {};
\node[fill=c1!15!c2] at (4em,6em) {};
\node[fill=c1!13!c2] at (5em,6em) {};
\node[fill=c1!11!c2] at (6em,6em) {};
\node[fill=c1!10!c2] at (7em,6em) {};
\node[fill=c1!8!c2] at (8em,6em) {};
\node[fill=c1!7!c2] at (9em,6em) {};
\node[fill=c1!0!c2] at (0em,5em) {};
\node[fill=c1!15!c2] at (1em,5em) {};
\node[fill=c1!18!c2] at (2em,5em) {};
\node[fill=c1!18!c2] at (3em,5em) {};
\node[fill=c1!16!c2] at (4em,5em) {};
\node[fill=c1!14!c2] at (5em,5em) {};
\node[fill=c1!12!c2] at (6em,5em) {};
\node[fill=c1!11!c2] at (7em,5em) {};
\node[fill=c1!10!c2] at (8em,5em) {};
\node[fill=c1!8!c2] at (9em,5em) {};
\node[fill=c1!0!c2] at (0em,4em) {};
\node[fill=c1!11!c2] at (1em,4em) {};
\node[fill=c1!19!c2] at (2em,4em) {};
\node[fill=c1!18!c2] at (3em,4em) {};
\node[fill=c1!18!c2] at (4em,4em) {};
\node[fill=c1!17!c2] at (5em,4em) {};
\node[fill=c1!15!c2] at (6em,4em) {};
\node[fill=c1!14!c2] at (7em,4em) {};
\node[fill=c1!12!c2] at (8em,4em) {};
\node[fill=c1!12!c2] at (9em,4em) {};
\node[fill=c1!0!c2] at (0em,3em) {};
\node[fill=c1!0!c2] at (1em,3em) {};
\node[fill=c1!15!c2] at (2em,3em) {};
\node[fill=c1!19!c2] at (3em,3em) {};
\node[fill=c1!20!c2] at (4em,3em) {};
\node[fill=c1!20!c2] at (5em,3em) {};
\node[fill=c1!19!c2] at (6em,3em) {};
\node[fill=c1!18!c2] at (7em,3em) {};
\node[fill=c1!17!c2] at (8em,3em) {};
\node[fill=c1!16!c2] at (9em,3em) {};
\node[fill=c1!0!c2] at (0em,2em) {};
\node[fill=c1!0!c2] at (1em,2em) {};
\node[fill=c1!8!c2] at (2em,2em) {};
\node[fill=c1!17!c2] at (3em,2em) {};
\node[fill=c1!20!c2] at (4em,2em) {};
\node[fill=c1!20!c2] at (5em,2em) {};
\node[fill=c1!21!c2] at (6em,2em) {};
\node[fill=c1!20!c2] at (7em,2em) {};
\node[fill=c1!19!c2] at (8em,2em) {};
\node[fill=c1!19!c2] at (9em,2em) {};
\node[fill=c1!0!c2] at (0em,1em) {};
\node[fill=c1!0!c2] at (1em,1em) {};
\node[fill=c1!0!c2] at (2em,1em) {};
\node[fill=c1!3!c2] at (3em,1em) {};
\node[fill=c1!12!c2] at (4em,1em) {};
\node[fill=c1!17!c2] at (5em,1em) {};
\node[fill=c1!19!c2] at (6em,1em) {};
\node[fill=c1!20!c2] at (7em,1em) {};
\node[fill=c1!21!c2] at (8em,1em) {};
\node[fill=c1!21!c2] at (9em,1em) {};
\node[fill=c1!0!c2] at (0em,0em) {};
\node[fill=c1!0!c2] at (1em,0em) {};
\node[fill=c1!0!c2] at (2em,0em) {};
\node[fill=c1!0!c2] at (3em,0em) {};
\node[fill=c1!0!c2] at (4em,0em) {};
\node[fill=c1!8!c2] at (5em,0em) {};
\node[fill=c1!13!c2] at (6em,0em) {};
\node[fill=c1!16!c2] at (7em,0em) {};
\node[fill=c1!19!c2] at (8em,0em) {};
\node[fill=c1!20!c2] at (9em,0em) {};
\tikzstyle{every node} = [];
\node at (4.5em,10.5em) {$\alpha$=5};
\node at (0em,-1em) {\tiny{20}};
\node at (1em,-1em) {\tiny{40}};
\node at (2em,-1em) {\tiny{60}};
\node at (3em,-1em) {\tiny{80}};
\node at (4em,-1em) {\tiny{100}};
\node at (5em,-1em) {\tiny{120}};
\node at (6em,-1em) {\tiny{140}};
\node at (7em,-1em) {\tiny{160}};
\node at (8em,-1em) {\tiny{180}};
\node at (9em,-1em) {\tiny{200}};

\end{scope}

\begin{scope}[xshift=12cm]
%\tikzstyle{every node} = [draw=none, rectangle, minimum width=1em, minimum height=1em];
%\node[fill=c1!90!c2] at (0em,9em) {};
%\node[fill=c1!80!c2] at (0em,8em) {};
%\node[fill=c1!70!c2] at (0em,7em) {};
%\node[fill=c1!60!c2] at (0em,6em) {};
%\node[fill=c1!50!c2] at (0em,5em) {};
%\node[fill=c1!40!c2] at (0em,4em) {};
%\node[fill=c1!30!c2] at (0em,3em) {};
%\node[fill=c1!20!c2] at (0em,2em) {};
%\node[fill=c1!10!c2] at (0em,1em) {};
%\node[fill=c1!0!c2] at (0em,0em) {};
\tikzstyle{every node} = [];
\node at (1.2em,9em) {\tiny{16\%}};
\node at (1.2em,0em) {\tiny{0\%}};
\node at (1.2em,4.5em) {\tiny{8\%}};
\draw[shading = axis, bottom color=c1!0!c2, top color=c1!100!c2, anchor=north] (-0.5em,-0.5em) rectangle (0.5em,9.5em);
\end{scope}

\end{tikzpicture}
\end{center}
\caption{This figure illustrates the effect of the problem settings (labeling budget, samples set, and imbalance severity) on the effectiveness of the active selection with the LSS criterion compared to a random selection. We report the difference in percentages of weighted accuracy between active and random sampling in the form of a heatmap. The heatmaps from left to right represent, respectively, highest to lowest imbalance severity (we vary the $\alpha$ of the Dirichlet distribution). The Y axis denotes the number of labels and the X axis the total number of samples. Note that for cases where the total number of samples is lower than that of the labels, we report a 0 difference in the two approaches.}
\label{heatmap}
\end{figure}
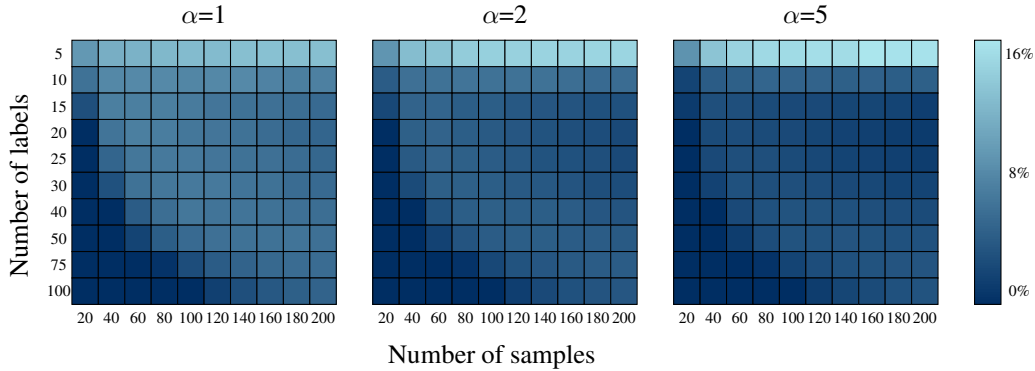

\subsection{Additional experiments}

We investigate further the effects of the active criterion in Table~\ref{ALC}. For each of the four datasets, we compare the sampling criteria performance to that of a random selection, and also to an oracle in the 5-labels case.
We compute the oracle by running a Nearest Class Mean on all the possible combinations of 5 labels from the 80 total samples. For this oracle, we only compute 1000 random runs for mini-ImageNet and 100 runs for FC-100, tiered-ImageNet and CUB (because of computational limitations as there are $\binom{80}{5}=24,040,016$ combinations). This highlights the potential performance that can be reached, especially as we already obtain decent scores comparable to the state-of-the-art standard configuration with simple heuristics. We can see that the margin and K-medoid perform well beyond the random selection and that the proposed LSS method slightly outperforms those criteria.

\begin{table}
\small
\centering
\setlength{\extrarowheight}{0pt}
\addtolength{\extrarowheight}{\aboverulesep}
\addtolength{\extrarowheight}{\belowrulesep}
\setlength{\aboverulesep}{0pt}
\setlength{\belowrulesep}{0pt}
\caption{Comparison of active selection criteria on all four considered datasets. We add an oracle strategy which consists of the selection that yields the best performance. We also chose not to include results for 25-labels for CUB for the same reasons mentioned in Table~\ref{soa_bis}. All values are percentages.}
\label{ALC}
\begin{tabular}{llllll} 
\midrule
\textbf{labels $\ell$}     & \textbf{sampling strategy}                           & mini-ImageNet                                             & tiered-ImageNet                                           & CUB                                                       & FC-100                                                     \\ 
\midrule
\multirow{5}{*}{5 labels}  & random                                               & 62.2 $\pm$ 0.31                                           & 61.2 $\pm$ 0.30                                           & 67.7 $\pm$ 0.33                                           & 46.5 $\pm$ 0.26                                            \\
                           & oracle                                               & 95.2 $\pm$ 0.21                                           & 97.4 $\pm$ 0.3                                            & 95.5 $\pm$ 0.18                                           & 80.2 $\pm$ 1.40                                            \\
                           & margin                                               & 69.0 $\pm$ 0.30                                           & 69.0 $\pm$ 0.26                                           & 78.5 $\pm$ 0.30                                           & 48.7 $\pm$ 0.26                                            \\
                           & K-medoid                                             & 75.6 $\pm$ 0.28                                           & 75.2 $\pm$ 0.27                                           & 83.8 $\pm$ 0.25                                           & 52.1 $\pm$ 0.25                                            \\
                           & {\cellcolor[rgb]{0.90,0.90,0.90}}\textbf{LSS (ours)} & {\cellcolor[rgb]{0.90,0.90,0.90}}\textbf{76.1 $\pm$ 0.26} & {\cellcolor[rgb]{0.90,0.90,0.90}}\textbf{75.7 $\pm$ 0.27} & {\cellcolor[rgb]{0.90,0.90,0.90}}\textbf{84.2 $\pm$ 0.24} & {\cellcolor[rgb]{0.90,0.90,0.90}}\textbf{53.3 $\pm$ 0.24}  \\ 
\midrule
\multirow{4}{*}{25 labels} & random                                               & 86.8 $\pm$ 0.18                                           & 85.6 $\pm$ 0.20                                           & \multicolumn{1}{c}{-}                                     & 68.8 $\pm$ 0.20                                            \\
                           & margin                                               & 88.6 $\pm$ 0.19                                           & 87.0 $\pm$ 0.18                                           & \multicolumn{1}{c}{-}                                     & 69.7 $\pm$ 0.21                                            \\
                           & K-medoid                                             & 89.7 $\pm$ 0.17                                           & 88.6 $\pm$ 0.16                                           & \multicolumn{1}{c}{-}                                     & 71.2 $\pm$ 0.19                                            \\
                           & {\cellcolor[rgb]{0.90,0.90,0.90}}\textbf{LSS (ours)} & {\cellcolor[rgb]{0.90,0.90,0.90}}\textbf{90.9 $\pm$ 0.16} & {\cellcolor[rgb]{0.90,0.90,0.90}}\textbf{89.3 $\pm$ 0.19} & \multicolumn{1}{c}{{\cellcolor[rgb]{0.90,0.90,0.90}}-} & {\cellcolor[rgb]{0.90,0.90,0.90}}\textbf{71.7 $\pm$ 0.20}  \\
\hline
\end{tabular}
\end{table}

Next, in Figure~\ref{heatmap} we challenge the effectiveness of active sampling over random sampling in different settings. We vary the total number of samples, the number of labels and finally the imbalance severity through the $\alpha$ parameter of the Dirichlet distribution. We observe smoother differences in the high severity cases, i.e a smaller range of values. Whereas in the softer imbalance case, we notice more clearly that active selection yields significantly better results for low labeling budgets and slightly better results for higher labeling budgets. This is the same effect as the one we briefly mentioned for Table~\ref{tab:table1_soa}: the active versus random gap closes as the labeling budget covers a larger proportion of the samples. We conclude that regardless of whether we have a more acute imbalance ($\alpha=1$) or a softer one ($\alpha=5$) the boost in performance remains consequent. We further investigate how class balance affects these findings for different datasets in the following section.

% \usepackage{multirow}
% \usepackage{multirow}

\subsection{Ablation study}
\label{ablation}

Finally, we investigate the effect of graph smoothing (Section~\ref{prepro}) on our method's efficacy. We also want to verify that the active selection methods remain relevant when the label distribution is balanced (i.e uniform) in the samples, as some studies (in different domains) found active methods to be less effective when it comes to balanced sets \cite{muller2022active}.

\begin{figure}[h!]
\begin{center}
\pgfplotsset{compat=1.7}

\pgfplotstableread[col sep = comma, header=true] {dat.csv}\mydata
\begin{tikzpicture}
\begin{axis}[legend style={nodes={scale=0.5, transform shape}}, 
        legend image post style={mark=*} ,title=mini-ImageNet, xlabel=queries, ylabel=accuracy, height=4.1cm, width=7cm , ymin=50,ymax=100,ytick={60,80},yticklabels={{60\%},{80\%}}]
\addplot [red,thick] table [x=a, y=b] \mydata;
\addlegendentry{LSS}
\addplot [blue,thick] table [x=a, y=c] \mydata;
\addlegendentry{random}
\addplot [red,dashed,thick] table [x=a, y=d] \mydata;
\addlegendentry{LSS no smoothing}
\addplot [blue,dashed,thick] table [x=a, y=e] \mydata;
\addlegendentry{random no smoothing}
\end{axis}
\begin{scope}[xshift=6.5cm]
\begin{axis}[legend style={nodes={scale=0.5, transform shape}}, 
        legend image post style={mark=*}, title=tiered-ImageNet,xlabel=queries, height=4.1cm, width=7cm, yticklabels={}, ymin=50,ymax=100]
\addplot [red, thick] table [x=a, y=f] \mydata;
\addlegendentry{LSS}
\addplot [blue, thick] table [x=a, y=g] \mydata;
\addlegendentry{random}
\addplot [red,dashed, thick] table [x=a, y=h] \mydata;
\addlegendentry{LSS no smoothing}
\addplot [blue,dashed, thick] table [x=a, y=i] \mydata;
\addlegendentry{random no smoothing}
\end{axis}
\end{scope}
\end{tikzpicture}
\end{center}
\caption{This figure showcases the impact of the smoothing in the performance of random and active selection on mini-ImageNet and tiered-Imagenet. The red lines correspond the LSS active method and the blue lines to the random selection. We use continuous lines for when graph smoothing is used and dashed lines for when graph smoothing is not used.}
\label{diffabl}
\end{figure}
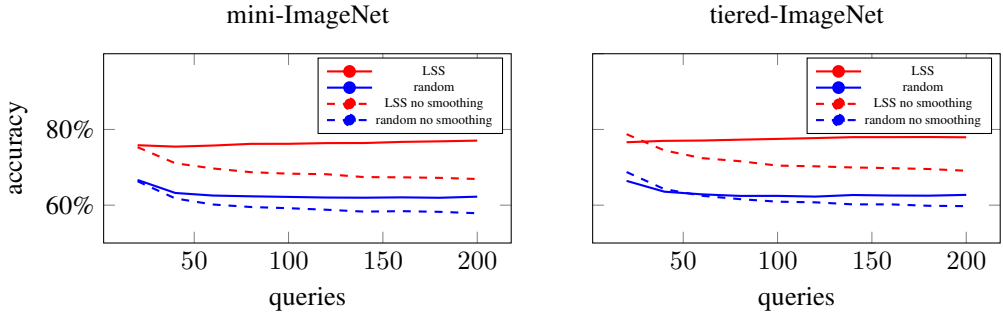

\textbf{Graph smoothing}: With graph smoothing applied to the data, we see a significant gap between the soft $K$-means performance in the random and active settings. In this section, we want to see how this gain from random to active is affected when we do not apply smoothing.
Figure~\ref{diffabl} shows the evolution of the weighted accuracies as we vary the number of samples. The goal is to capture the effect of smoothing on different graph sizes. We see similar effects in both mini-ImageNet and tiered-Imagenet. We observe that the smoothing process distinctively benefits the performance of the active selection especially for bigger samples pools. We also note that it increases, for all sample sizes, the gap between the performance of the random and active selection.

\textbf{Data Imbalance}: We reproduce our experiments on mini-ImageNet, tiered-ImageNet and FC-100 in a balanced class setting. Table~\ref{balanced} shows the results for a random sampling, and an active sampling with our best criterion in the balanced class scenario. Compared to the imbalanced setting, we see an expected increase in performance for all methods across all datasets. But most importantly, we observe that the active sampling outperforms the random selection for both the 5-labels and 25-labels cases with comparable margins to the ones in the balanced scenario (Table~\ref{ALC}). We see a larger gap in the 5 labels setting and notice that active sampling, in this setting, benefits more the accuracies when classes are balanced whereas in the 25 labels scenario the active vs random gaps are smaller and less consequent than those in the imbalanced scenario. This further proves that the active sampling is overall much more effective than its random counterpart regardless of class label distribution.

\begin{table}
\centering
\caption{Active sampling versus random sampling for balanced samples set scenarios.
We consider the 5 and 25 labels cases for a total sample set of size 80 and 100 respectively. This table also includes a row for the Gap (LSS-Random) in this balanced setting, and for the gap differences between the gap in the balanced classes and the gap in the imbalanced classes with $\alpha$=2: Gap diff = Gap imbalanced - Gap balanced. All values are percentages.}
\small
\label{balanced}
\begin{tabular}{lllll} 
%\cmidrule(l){3-5}
%                                    &                                   & \multicolumn{3}{c}{\textbf{Datasets}}            \\ 
\midrule
\textbf{labels $\ell$ number}        & \textbf{sampling strategy}        & mini-Imagenet & tiered-Imagenet & FC-100         \\ 
\midrule
\multirow{4}{*}{\textbf{5 labels}}  & random                            & 67.4 $\pm$ 0.30 & 64.3  $\pm$ 0.32 & 49.0 $\pm$ 0.26    \\
                                    & \textbf{LSS}                      & \textbf{82.9 $\pm$ 0.21} & \textbf{81.3 $\pm$ 0.23}   & \textbf{56.9 $\pm$ 0.23}  \\
                                    & Gap (LSS-random)                  & 15.5          & 17            & 7.9            \\
                                    & \textbf{Gap diff (with imbalanced)}          & \textbf{-2.7} & \textbf{-3.3}   & \textbf{-1.1}  \\ 
\hline
\multirow{4}{*}{\textbf{25 labels}} & random                            & 90.1  $\pm$ 0.11 & 88.6  $\pm$ 0.16  & 71.3 $\pm$ 0.17 \\
                                    & \textbf{LSS}                      & \textbf{92.4 $\pm$ 0.12} & \textbf{90.2 $\pm$ 0.16}   & \textbf{72.9 $\pm$ 0.18}  \\
                                    & Gap (LSS-random)                  & 2.3           & 1.6               & 1.6            \\
                                    & \textbf{Gap diff (with imbalanced)} & \textbf{1.5}  & \textbf{2.1}    & \textbf{1}   \\
\hline
\end{tabular}
\end{table}

\subsection{Limitations}
\label{lim}

While we do observe very encouraging results that prove the utility of opting for active selection methods, it remains important to note that this proposed framework could be further explored and challenged in a number of other regards. A notable one is our usage of the class mean of the labels in the statistical inference for the soft K-means initialization. We suspect that this choice could have intriguing effects on the performance of the different sampling criteria through the loss of information resulting from the averaging operation. Also, we suppose known and given the number of classes $K$ in the considered task, which could be unrealistic in some applications.

\section{Related work}
\label{rw}

\textbf{Few-Shot Learning}: The Few-Shot learning problem is one that has been attracting more and more attention over the last couple of years. Different branches are being explored and major ones include meta-learning, which aims to acquire task-level meta knowledge that the model utilizes to quickly adapt to new tasks with very few labeled examples \cite{vilalta2002perspective,rusu2018meta,nichol2018first,finn2017model}. Another major branch is metric-learning, where the model leverages distance metrics to estimate the similarity between two inputs. Learning this allows generalization to novel categories with few labeled instances, \cite{yang2006distance,kulis2013metric,koestinger2012large}. These methods require for a model to be pre-trained on more general tasks, and the advances in this field made available a myriad of efficiently trained feature extractors with data augmentation techniques \cite{hu2021leveraging,simon2020adaptive,li2020adversarial}, ensembling \cite{bendou2022easy} and various other techniques. Classification in the inductive setting has mostly relied on simple methods \cite{bendou2022easy} whereas more diverse methods were used in the transductive setting allowing for more promising results \cite{hu2021leveraging,ziko2020laplacian,liu2020prototype,boudiaf2020information}. The authors of \cite{veilleux2021realistic} have also challenged class balance priors in this setting and its importance within the recent line of works. Our work falls within this rethinking of the transductive configuration towards a more class and label distribution agnostic approach.
Our work also relates to Self Supervised Techniques \cite{chen2021self,yue2021prototypical}, as the Cold-start aspect i.e starting from zero examples is pivotal for us and we use unsupervised methods for as long as we do not have any labels readily available.

\textbf{Active learning}: A core part of this paper revolves around the significant increase in scores that we can see if we were to select the right labels, and around an exploration of the heuristics that we considered to complete this selection. A lot of the current literature in Active Learning remains unsuited for Few-Shot scenarios as most used techniques that have strong theoretical bases are often heavily reliant on the utilized models. These techniques usually seek out to decrease the learner's variance, or base the selection on the uncertainty of the model \cite{cohn1996active,settles2009active,gal2017deep,wang2016cost,zhang2022galaxy}. These techniques become challenging in a Cold-start \cite{houlsby2014cold,yuan2020cold} setting where we start from zero labeled examples. Having too few samples and/or labeling budget makes this even more challenging as we cannot afford to undergo an initial phase of irrelevant labelings based on an unstable model. However, some recent papers showed that in Natural Language Processing the pre-trained models can be used to overcome this challenge \cite{muller2022active}. In our few shot configuration, we also leverage the feature extractors to obtain "nicer" data distributions where we could mix diversity sampling methods and uncertainty based sampling methods \cite{yang2015multi,wang2017uncertainty,smith2018less}to obtain an effective labeling pipeline.

\section{Conclusion}

We introduced a novel formulation for the problem of Active Few-Shot Classification (AFSC). This problem can be seen as a new paradigm rival to Transductive Few-Shot Classification (TFSC). We proposed a simple methodology that relies on two main steps: a statistical inference method and an active sampling step in which we select samples to label based on Log-probs Soft K-Means Sampling. We adapted benchmarks from the field of TFSC and we proved that the proposed framework can achieve new levels in accuracy compared to TFSC. We expect AFSC to be applicable to a wide range of real-world applications, and as such to define a new promising framework for future contributions.

\section{Acknowledgements}

This work has been performed as part of a CIFRE PhD program supported by Schneider Electric and ANRT (Association Nationale Recherche Technologie).

{\small
\bibliographystyle{unsrt}
\bibliography{refs}
}

%%%%%%%%%%%%%%%%%%%%%%%%%%%%%%%%%%%%%%%%%%%%%%%%%%%%%%%%%%%%

%%% BEGIN INSTRUCTIONS %%%
%The checklist follows the references.  Please
%read the checklist guidelines carefully for information on how to answer these
%questions.  For each question, change the default \answerTODO{} to \answerYes{},
%\answerNo{}, or \answerNA{}.  You are strongly encouraged to include a {\bf
%justification to your answer}, either by referencing the appropriate section of
%your paper or providing a brief inline description.  For example:
%\begin{itemize}
  %\item Did you include the license to the code and datasets? \answerYes{See Section~\ref{gen_inst}.}
  %\item Did you include the license to the code and datasets? \answerNo{The code and the data are proprietary.}
  %\item Did you include the license to the code and datasets? \answerNA{}
%\end{itemize}
%Please do not modify the questions and only use the provided macros for your
%answers.  Note that the Checklist section does not count towards the page
%limit.  In your paper, please delete this instructions block and only keep the
%Checklist section heading above along with the questions/answers below.
%%% END INSTRUCTIONS %%%

%%%%%%%%%%%%%%%%%%%%%%%%%%%%%%%%%%%%%%%%%%%%%%%%%%%%%%%%%%%%

%\appendix

%\section{Appendix}

%Optionally include extra information (complete proofs, additional experiments and plots) in the appendix.
%This section will often be part of the supplemental material.

\end{document}